# Distilling Neuron Spike with High Temperature in Reinforcement Learning Agents

Ling Zhang[1], Jian Cao[1]*, Yuan Zhang[1], Bohan Zhou[1], Shuo Feng[1]
1 School of Software and Microelectronics, Peking University, Beijing, China.
{zhangling@stu.pku.edu.cn, caojian@ss.pku.edu.cn, zhangyuan@stu.pku.edu.cn}

**Abstract**

Spiking neural network (SNN), compared with depth neural network (DNN), has faster processing speed, lower energy consumption and more biological interpretability, which is expected to approach Strong AI. Reinforcement learning is similar to learning in biology. It is of great significance to study the combination of SNN and RL. We propose the reinforcement learning method of spike distillation network (SDN) with STBP. This method uses distillation to effectively avoid the weakness of STBP, which can achieve SOTA performance in classification, and can obtain a smaller, faster convergence and lower power consumption SNN reinforcement learning model. Experiments show that our method can converge faster than traditional SNN reinforcement learning and DNN reinforcement learning methods, about 1000 epochs faster, and obtain SNN 200 times smaller than DNN. We also deploy SDN to the PKU nc64c chip, which proves that SDN has lower power consumption than DNN, and the power consumption of SDN is more than 600 times lower than DNN on large-scale devices. SDN provides a new way of SNN reinforcement learning, and can achieve SOTA performance, which proves the possibility of further development of SNN reinforcement learning.

## Introduction

Brain is the most important organ of animals. Since Cesar Julien Jean legallois defined the specific functions of brain regions [1], brain neurology has attracted many researchers, and many experiments to reproduce the brain system have been carried out. As early as 1958, Rosenblatt built a perceptron model based on brain neurons [2], which opened the way for neural network (NN). Neuralink made an attempt to convert the activity of brain neurons into commands to control external devices in monkeys [3]. The neural pattern recognition machine architecture [4] built by Stephen Grossberg et al. also provided ideas for the later design of various neural morphology chips, including TrueNorth [5], Loihi [6] and Tianjic [7].

Since AlexNet [8], deep neural network (DNN) has shown superior performance in speech and image processing. However, the high accuracy and performance of DNN are based on very high power consumption. In addition, the development of DNN has gradually deviated from its original intention of imitating human brain and creating intelligent life. Spiking neural network (SNN), as the third generation neural network, has lower power consumption and faster speed than DNN, and is also more biologically interpretable [9]. It is expected to complete more complex cognitive problems in the form closest to the brain neuron paradigm and create the so-called "electronic life". On the other side, the implementation and application of SNN is a bold trial of brain system modeling, which may provide precious and meaningful experience for neurology.

SNN based applications are gradually improving: in terms of classification, Riccardo Massa et al. use DVS for gesture recognition [10]; In terms of object detection, Kim S et al. proposed spiking Yolo [11]. The above supervised learning applications have achieved good results in accuracy by learning labels. However, in order to achieve strong AI, unsupervised learning and reinforcement learning are indispensable. Reinforcement learning is a way for agents to continuously learn through interaction with the environment and reward and punishment mechanism. This way is very similar to the growth mode of organisms and coincides with the research concept of SNN.

Research on SNN reinforcement learning is also ongoing: Nan Zheng et al. trained the hardware friendly actor critical network by using spiking-timing depedent plasticity rule (STDP) [12]. Mengwen Yuan et al. trained the SDC network by combining the hedonistic synapse model and STDP [13]. Vahid Azimirad et al. trained the TCT network by using STDP [14]. Fatemeh Sharifizadeh et al. trained r-snn network using r-stdp [15]. Devdhar Patel et al. trained DQN network based on DNN, and then converted it to SNN [16]. Most studies tend to use STDP and its improved version to complete SNN reinforcement learning. At the same time, some studies look for new solutions.

In 2018, Yujie Wu et al. proposed the spatio-temporal backpropagation (STBP) algorithm for training high-performance SNN [17]. Experiments showed that the performance of SNN trained by STBP algorithm is much higher than that of STDP algorithm, which is a new way to train SNN. However, STBP algorithm uses spike-rate coding [18], which will extremely compress the reinforcement learning action value's search space, resulting in the difficulty of using STBP algorithm directly for training action value based SNN reinforcement learning. Guangzhi Tang et al. implemented the network based on action value in DDPG with DNN and the network based on deterministic strategy with SNN to avoid the operation of action value search by SNN[19]. This is one of the few studies research using STBP algorithm to train SNN reinforcement learning model. This paper proposes another method to realize the action value based SNN, and is expected to use this method to train smaller SNN.

In this paper, we propose the SNN reinforcement learning method of Spiking Distillation Network (SDN) with STBP, which has achieved better results than the traditional SNN reinforcement learning and DNN reinforcement learning algorithms. We also give the proof that spike-rate coding in STBP will extremely compress the reinforcement learning action value's search space. We got inspiration from knowledge distillation [20] and transformed the original way of DNN teacher network training DNN student network into DNN teacher network training SNN student network. In this way, DNN will carry out reinforcement learning training and search action space. Then SNN will learn from DNN. This effectively avoids the work of SNN searching action space, and can further compress the capacity of SNN.

In the following, this paper first introduces the prior knowledge, then introduce our SDN method in detail, and proves the disadvantage of action value search for STBP through the formula. Then, we compare SDN with other SNN and DNN reinforcement learning methods to prove the effectiveness and efficiency of SDN, and prove the compressibility of SDN by comparing the capacity of teacher network and student network in SDN. In addition, we also deploy the SDN model on PKU NC64C chip to prove the low power consumption of SDN.

## METHODS AND MATERIALS

We focus on how to effectively use STBP to train a more intelligent SNN reinforcement learning model, while avoiding the problem of STBP compressing the action search space. In this section, we introduce our method in detail. At the same time, we prove that STBP will greatly compress the reinforcement learning action search space to illustrate the necessity of our method.

### Training of Teacher Net

The training of Spiking Distillation Network (SDN) is based on a well-trained DNN reinforcement learning teacher network. We use DQN [21] to train out DNN teacher network.

We consider a scene where an agent interacts with the environment. The environment provides the agent with T previous states from the current moment, $S \in R^{T \times H \times W}$, and reward at each time. The state provided by the environment is the image, H and W represent the height and width of the image, and the agent chooses the next action $a_1$ or $a_2$ according to state. The agent inputs the obtained state into DNN. DNN predicts the Q value of $a_1$ and $a_2$, which represents the accumulation of reward that can be obtained from the current moment to the end of the game. The agent chooses the action with the highest Q value as its next instant action.

For DQN, the loss function is as follows:

$$L = \frac{1}{n}\sum_{i=1}^{n}(R_t + \gamma max(Q^L(X_{t+1})) - max(Q^L(X_t)))^2 \quad (1)$$

Where $Q^L$ is the DNN of layer L, $\gamma$ is the suspect value of DNN prediction result, $X_t$ and $R_t$ are randomly taken from the experience pool, which stores the environment state and reward within a period of time.

It should be noted that SDN is not limited to the training methods of reinforcement learning DNN. We can use any deep reinforcement learning methods to train our DNN teacher network, such as DDPG[22], TD3[23], PPO[24], etc., and then use SDN to train SNN.

### Architecture of Spiking Distillation Network

Spiking Distillation Network (SDN) is a training framework that uses DNN teachers to guide SNN students. Compared with training of teacher network, the environment provides different preprocessed states to DNN teacher network and SNN student network respectively, and does not provide reward.

For DNN teacher network, the environment provides it with T previous states from the current moment, $S \in R^{T \times H \times W}$. For SNN student network, the environment does not provide it with T previous states at the same time. Instead, these states, $s_t \in R^{H \times W}$, are provided in chronological order, which will provide SNN with strong time dimension information.

We use leaky integrate-and-fire (LIF) neurons to describe our SNN model. The state of LIF neurons is determined by spike input, membrane potential and threshold potential. The membrane potential accumulates with the spike input. When the membrane potential exceeds the threshold potential, LIF neurons fire spikes. The membrane

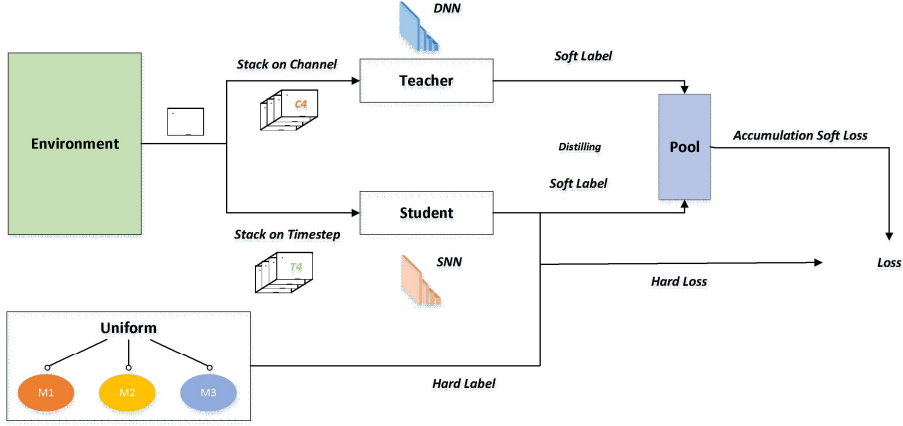

Fig 1. Flow chart of SDN algorithm

potential will leak and decrease over time. The differential expression of LIF neurons is as follows:

$$\tau \frac{du(t)}{dt} = -u(t) + i(t) \quad (2)$$

Where $u(t)$ is the membrane potential of LIF neuron at time t, $\tau$ is leakage rate, and $I(t)$ is the product of input spike and weight.

In order to use STBP, we need to divide the expression of LIF neuron into three parts: Calculation of $I(t)$, accumulation of membrane potential $u(t)$ and firing spike, which is described in algorithm 1. Where $t$ is the time, $l_1$ is the number of SNN student network's layers, $o_t^{l_1}$ represents the spike firing's situation of SNN layer L at time t, and *thresh* is the threshold potential, $sp$ represents the number of output spikes, and $fr$ represents the output with spike-rate coding.

---

Algorithm 1: Accumulation of experience pool for SDN

**Input**: Environment state $S \in R^{T \times H \times W}$ and $s_t \in R^{H \times W}$
**Parameter**: weights $W$ and biases $b$ of SNN student network
**Output**: firing rates $fr$ of output
1: Environment provides state $S$ and $s_t$
2: store $S$ and $s_t$ to pool $D$
3: **while** $t = 1,\ldots,T$ **do**
4:   $s_t$ input to SNN student network
5:   **while** $l_1 = 1,\ldots,L_1$ **do**
6:     update LIF neurons:
7:     $i_t^{l_1} = W^{l_1} o_t^{l_1-1} + b^{l_1}$
8:     $u_t^{l_1} = \tau u_{t-1}^{l_1} \cdot (1 - o_{t-1}^{l_1}) + i_t^{l_1} + b_t^{l_1}$
9:     $o_t^{l_1} = thresh(u_t^{l_1})$
10:   **end while**
11:   Accumulating the output spikes of student SNN:
12:   $sp += o_t^{L_1}$
13: **end while**
14: Compute firing rates of output: $fr = sp / T$
15: Provide the environment with $fr$

---

$\tau$ is an important parameter used by LIF neurons in STBP to obtain time dimension information, which can distinguish input in different time sequences. Assuming that $u_t$ has not reached the threshold potential from time 0 to time t, the membrane potential of LIF neurons at time t is as follows:

$$u_T^{l_1} = \tau^{T-1} u_1^{l_1} + \tau^{T-2} u_2^{l_1} + \ldots + \tau^2 u_{T-2}^{l_1} + \tau u_{T-1}^{l_1} + i_T^{l_1} + b_T^{l_1} \quad (3)$$

It can be seen that $\tau$ completes the distinction of input in different time sequences through its own cumulative multiplication.

The convolution of DNN is difficult to make a strong distinction between the input in different time sequences on different channels, because the pixels at the same position on all channels will be uniformly multiplied and added to obtain an output pixel, as shown in the following formula:

$$a_{t,x_o,y_o} = \sum_{t=1}^{T} \sum_{y_i=y_o}^{y_o+k} \sum_{x_i=x_o}^{x_o+k} b + w_t s_{(t,x_i,y_i)} \quad (4)$$

Where $a_{t,x_o,y_o}$ represents the pixel at position $x_o, y_o$ and channel $t$ in the first layer of DNN, $S_{t,x_i,y_i}$ represents the pixel at position $x_i, y_i$ and channel $t$ in the input, and $w_t$ represents the weight in DNN.

This means that when the input with different time order on the channel is sent into the first layer of DNN, each channel output by the first layer will no longer have a strong time order difference, although each pixel of the output is fused with the information of different time dimensions, which we called weak difference.

SDN will first accumulate experience pool, saving $S$ and $S_t$ given by the environment. $S_t$ is sent to the SNN student network, obtains the output $fr$, and we provide $fr$ to the environment as the judgment basis for the next action selection. When a certain epoch is reached and the experience pool is full, SDN will begin to use the content stored in the experience pool to train the SNN student network. Among them, the role of experience pool is to disrupt the relevance of samples and improve the utilization

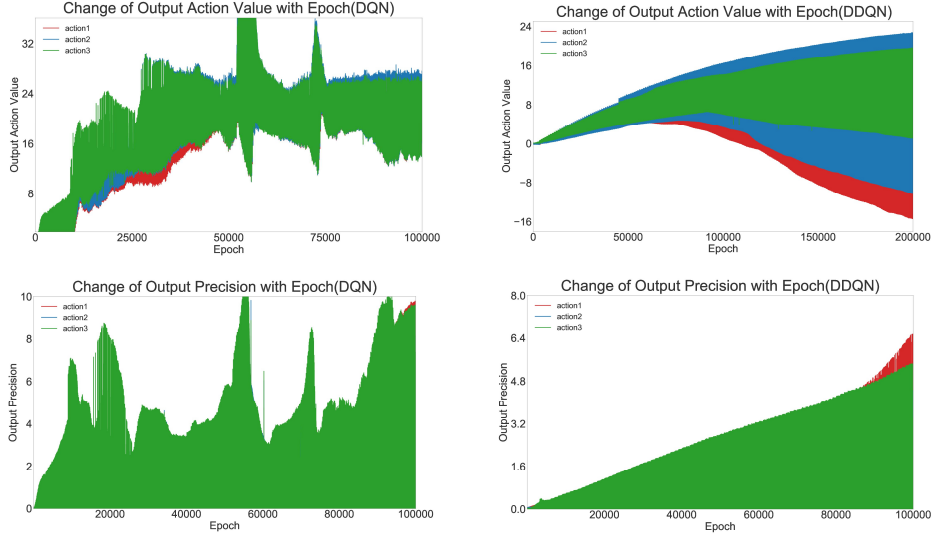

Fig 2. Exploration of action space in DNN reinforcement learning

of data. The overall architecture of SDN is shown in the figure below:

## Training of SDN

The training process of SDN is shown in algorithm 2. At the beginning, SDN accumulates experience pool. When epoch is greater than $thresh\_epoch$, the experience pool is full, and we will take out the corresponding environment state $S^b$ and $s_t^b$ from experience pool D in batches, input them into SNN student network and DNN teacher network respectively for forward propagation, obtain the corresponding output, $fr^b$ and $o^b$, and calculate the loss of these two outputs.

The loss function we use is CrossEntropy with T: we will divide the output by a temperature value to soften the output, and then calculate the SoftLoss through Softmax and CrossEntropy. At the same time, we will make the output of the teacher network one hot as a hard label, and then use this label to calculate the HardLoss through CrossEntropy with the output of the student network. Add Softloss and Hardloss in proportion to obtain the final loss function.

Then, STBP back propagation is performed on the student SNN using the calculated loss, and the parameters of the student SNN are updated from SD and TD.

## Impact of STBP spike-rate coding on action value exploration space

We have counted the change of the output and output precision of DNN based on DQN and DDQN [25], as shown in the figure below. It can be seen that the action value of DNN output is trained from a small negative number to a large positive number. However, the output range of spike-rate coding is only between 0 and 1, which is difficult to express the action value with such a large change range, which makes it very difficult to directly train SNN in DQN and DDQN based on action value reinforcement learning with STBP.

---

Algorithm 2: Training of SDN

**Input**: $S^b$ and $s_t^b$ randomly sampled in batches from experience pool $D$
**Parameter**: weights and biases of SNN student network and DNN teacher network
**Output**: Total Loss

1: **while** epoch = 0, …, epoch_end **do**
2:     Accumulation of experience pool for SDN
3:     **if** (epoch > thresh_epoch) **then**
4:         $s_t^b$ input to SNN Student network
5:         $fr^b = forward_{SNN}(s_t)$
6:         $S$ input to DNN Teacher network
7:         $o^b = forward_{DNN}(S)$
8:         LIF Loss compute:
9:         $q^b = exp(o^b/T)/\sum_j exp(o^b/T)$
10:       $z^b = exp(fr^b/T)/\sum_j exp(fr^b/T)$
11:       $l^b = onehot(o^b)$
12:       $SoftLoss = -\sum_b(z^b log(q^b) + (1-z^b)log(1-q^b/T))$
13:       $HardLoss = -\sum_b(l^b log(fr^b) + (1-l^b)log(1-fr^b))$
14:       $Loss = \lambda \cdot SoftLoss + (1-\lambda) \cdot HardLoss$
15:     STBP backward with loss in Student SNN
16:     **end if**
17: **end while**

We consider using affine transformation to map the output $x$ of DNN training to the range of STBP spike-rate coding $x_{fr}$ with accuracy $acc_{fr}$.

$$z = round(\frac{x_{min}}{x_{max} - x_{min}} \times acc_{fr}) / acc_{fr} \quad (5)$$

Where

$$x_{fr} = round(\frac{x}{x_{max} - x_{min}} \times acc_{fr}) / acc_{fr} + z \quad (6)$$

In order that this affine transformation will not cause loss, we need to specify at least:

$$\frac{acc_x}{x_{max} - x_{min}} \times acc_{fr} = 1, \; step = \frac{acc_x}{x_{max} - x_{min}} \quad (7)$$

Combined with the above data, the step required for spike-rate coding will reach the order of 1e5, which means that it needs to consume a lot of energy and inference speed, which brings a great burden to inference and training. SDN effectively avoids such search and completes the reinforcement learning training of SNN with few step.

## Experiments

In the first part, we tried different Loss Functions for SDN and got the best Loss Function: CrossEntropy with Temperature(T). Then, we compared our SDN with other SNN methods to confirm the effectiveness of SDN. In addition, we compared DNN reinforcement learning methods with SDN, and the result shows that SDN has a good learning ability. All experiments were performed in Pong game environment, as shown by Figure 3.

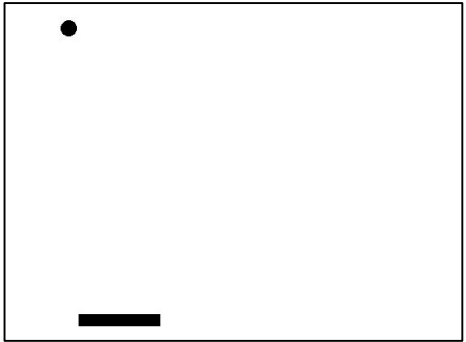

Fig 3. Pong Game

### Comparison of different Loss treatments of SDN

For the Loss Function in SDN, we tested the mean square error function and the cross-entropy function, as shown in Figure 4. For mean square error, since the range of DNN's output is not fit for the range of SNN output spike-rate coding, we tried to map the output of DNN into a range we set, so that the Loss Function can be calculated by the DNN and SNN output. Based on the degree of softness of DNN output, we divided the mapping range into three separate experiments: (1) Set the maximum value of DNN output to 1, and the other to 0. (2) Linear zoom the maximum value in DNN output to a range of 0 to 1, and the other to 0. (3) Linear scaling of the maximum value in DNN output (2) and adjust the remaining values in formula X. The adjustment in formula x is to widen the difference between the maximum and other values.

$$o_{max} = o_{max} / scale \quad (8)$$
$$o_{other} = o_{other} / scale - d \quad (9)$$

Where the scale is the scaling factor for linear scaling and D is used to widen the difference between the maximum and other values.

As for CrossEntropy with T, we set $T=10$, $\lambda=0.9$.

From the Fig x, reward in CrossEntropy with T training method is the most, so we chose CrossEntropy with T as the loss function in SDN.

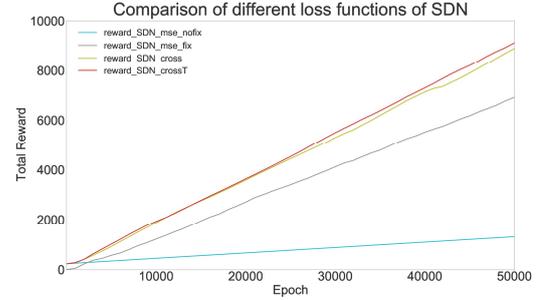

Fig 3. Comparison of different Loss treatments of SDN

### Comparison of SDN and other SNN reinforcement learning methods

We compared SDN with other SNN reinforcement learning methods, as shown in Figure x. It can be found that the STBP combined with DNN can not achieve an ideal learning effect, but the effect of SDN is much better than that of reinforcement learning with STBP, which is in line with the above explanation for impact of STBP spike-rate coding on action value exploration space.

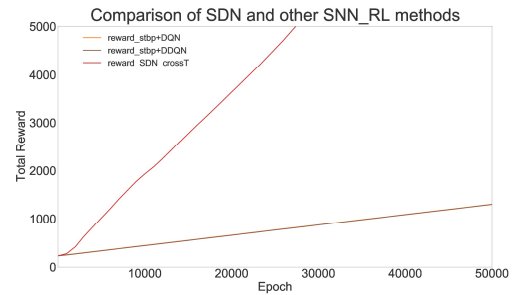

Fig 4. Comparison of SDN and other SNN reinforcement learning methods

## Comparison of teacher network and student network in SDN

We took DNN teacher network with larger structure to train smaller SNN student network. In pong game, we used 10 layers of teacher network with a size of 892 KB, while the trained SNN student network has only 3 layers and a size of 5 KB, which is nearly 200 times compressed, which proved that SDN has the ability to train smaller SNNs.

Tab1. Capacity of Teacher Network and Student Network

| Model | Layer nums | Capacity(KB) |
| --- | --- | --- |
| Teacher Network | 10 | 892 |
| Student Network | 3 | 5 |

## Comparison of SDN and DNN

Furthermore, we tried to compare the DNN reinforcement learning methods with SDN, as shown in Figure XX. Due to the combination of time dimension information, the learning speed of SDN is faster than that of DNN and the reward obtained is more than that of DNN.

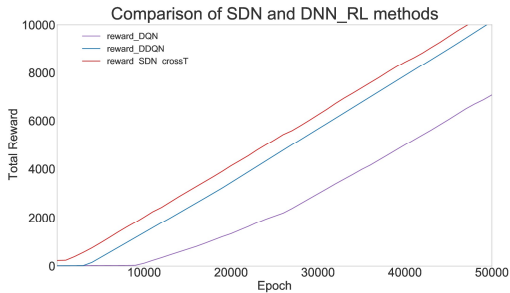

Fig 5. Comparison of SDN and DNN

## Deployment

In order to prove the effect of SDN in low power consumption, we deployed it on PKU NC64C chip [26] and compared it with various devices, as shown in tab.xx. It can be seen that compared with large-scale devices, such as NVIDIA rtx2080, the power consumption of SDN is reduced by more than 600 times; Compared with embedded devices, such as NVIDIA Jetson NX, SDN also consumes more than 10 times less power.

Tab2. Power performance of different devices

| Model | Device | Power(W) |
| --- | --- | --- |
| DNN | Nvidia RTX2080 | 64 |
| DNN | Nvidia RTX8000 | 62 |
| DNN | Nvidia Jetson NX | 1.429 |
| SDN | PKU NC64C | 0.103 |

## Discussion and Conclusion

Compared with DNN, SNN has lower power consumption and faster speed, and is closer to the form of brain neuron paradigm. Compared with other types of machine learning, reinforcement learning is closer to the essence of biological learning growth. Therefore, it is of great significance to explore the combination of the two.

This paper proposes a new SNN reinforcement learning algorithm: Spiking Distillation Network (SDN) with STBP. This method hands over the search of action value space to DNN, and makes DNN teacher network train SNN student network through similar methods of knowledge distillation. SDN can achieve better results than traditional SNN and DNN reinforcement learning algorithms.

We use the formula to prove that STBP is not conducive to the search of action value space, indicating the necessity of SDN. In addition, we proved through experiments that the convergence speed of SDN is faster than the traditional SNN and DNN reinforcement learning algorithms, which shows high efficiency of SDN. At the same time, we also proved through experiments that the capacity of student net in SDN can be much smaller than that of teacher net, which shows compressibility of SDN. Finally, we deploy SDN on PKU NC64C chip and the results show that its power consumption is less than other devices, which shows low power consumption of SDN.

For the future work, we can transplant more DNN reinforcement learning algorithms into SDN, such as A3C, DDPG, etc., try to use other distillation methods to improve the accuracy of SNN in SDN, and use new methods to conduct more game training. devices, such as NVIDIA Jetson NX, SDN also consumes less than 10 times power.

## Reference

[1] Finger S. Origins of neuroscience: a history of explorations into brain function[M]. Oxford University Press, USA, 2001.

[2] Rosenblatt F. The perceptron: a probabilistic model for information storage and organization in the brain[J]. Psychological review, 1958, 65(6): 386.

[3] Pisarchik A N, Maksimenko V A, Hramov A E. From novel technology to novel applications: Comment on "An integrated brain-machine interface platform with thousands of channels" by Elon Musk and Neuralink[J]. Journal of medical Internet research, 2019, 21(10): e16356.

[4] Carpenter G A, Grossberg S. A massively parallel architecture for a self-organizing neural pattern recognition machine[J].

Computer vision, graphics, and image processing, 1987, 37(1): 54-115.

[5] Akopyan F, Sawada J, Cassidy A, et al. Truenorth: Design and tool flow of a 65 mw 1 million neuron programmable neurosynaptic chip[J]. IEEE transactions on computer-aided design of integrated circuits and systems, 2015, 34(10): 1537-1557.

[6] Davies M, Srinivasa N, Lin T H, et al. Loihi: A neuromorphic manycore processor with on-chip learning[J]. Ieee Micro, 2018, 38(1): 82-99.

[7] Pei J, Deng L, Song S, et al. Towards artificial general intelligence with hybrid Tianjic chip architecture[J]. Nature, 2019, 572(7767): 106-111.

[8] Krizhevsky A, Sutskever I, Hinton G E. Imagenet classification with deep convolutional neural networks[J]. Advances in neural information processing systems, 2012, 25: 1097-1105.

[9] Ghosh-Dastidar S, Adeli H. Spiking neural networks[J]. International journal of neural systems, 2009, 19(04): 295-308.

[10] Massa R, Marchisio A, Martina M, et al. An efficient spiking neural network for recognizing gestures with a dvs camera on the loihi neuromorphic processor[J]. arXiv preprint arXiv:2006.09985, 2020.

[11] Kim S, Park S, Na B, et al. Spiking-YOLO: Spiking neural network for energy-efficient object detection[C]//Proceedings of the AAAI Conference on Artificial Intelligence. 2020, 34(07): 11270-11277.

[12] Zheng N, Mazumder P. Hardware-Friendly Actor-Critic Reinforcement Learning Through Modulation of Spiking-Timing Dependent Plasticity[J]. IEEE Transactions on Computers, 2017, 66(2).

[13] Yuan M, Wu X, Yan R, et al. Reinforcement learning in spiking neural networks with stochastic and deterministic synapses[J]. Neural computation, 2019, 31(12): 2368-2389.

[14] Azimirad V, Sani M F. Experimental study of reinforcement learning in mobile robots through spiking architecture of Thalamo-Cortico-Thalamic circuitry of mammalian brain[J]. Robotica, 2020, 38(9): 1558-1575.

[15] Azimirad V, Sani M F. Experimental study of reinforcement learning in mobile robots through spiking architecture of Thalamo-Cortico-Thalamic circuitry of mammalian brain[J]. Robotica, 2020, 38(9): 1558-1575.

[16] Patel D, Hazan H, Saunders D J, et al. Improved robustness of reinforcement learning policies upon conversion to spiking neuronal network platforms applied to Atari Breakout game[J]. Neural Networks, 2019, 120: 108-115.

[17] Wu Y, Deng L, Li G, et al. Spatio-temporal backpropagation fo r training high-performance spiking neural networks[J]. Frontiers in neuroscience, 2018, 12: 331.

[18] Gerstner W, Kistler W M. Spiking neuron models: Single neurons, populations, plasticity[M]. Cambridge university press, 2002.

[19] Tang G, Kumar N, Michmizos K P. Reinforcement co-learning of deep and spiking neural networks for energy-efficient mapless navigation with neuromorphic hardware[J]. arXiv preprint arXiv:2003.01157, 2020.

[20] Hinton G, Vinyals O, Dean J. Distilling the knowledge in a neural network[J]. arXiv preprint arXiv:1503.02531, 2015.

[21] Mnih V, Kavukcuoglu K, Silver D, et al. Playing atari with deep reinforcement learning[J]. arXiv preprint arXiv:1312.5602, 2013.

[22] Lillicrap T P, Hunt J J, Pritzel A, et al. Continuous control with deep reinforcement learning[J]. arXiv preprint arXiv:1509.02971, 2015.

[23] Fujimoto S, Hoof H, Meger D. Addressing function approximation error in actor-critic methods[C]//International Conference on Machine Learning. PMLR, 2018: 1587-1596.

[24] Schulman J, Wolski F, Dhariwal P, et al. Proximal policy optimization algorithms[J]. arXiv preprint arXiv:1707.06347, 2017.

[25] Van Hasselt H, Guez A, Silver D. Deep reinforcement learning with double q-learning[C]//Proceedings of the AAAI conference on artificial intelligence. 2016, 30(1).

[26] Kuang Y, Cui X, Zhong Y, et al. A 64K-neuron 64M-1b-synapse 2.64 pJ/SOP Neuromorphic Chip with All Memory on Chip for Spike-based Models in 65nm CMOS[J]. IEEE Transactions on Circuits and Systems II: Express Briefs, 2021.